\documentclass[10pt,twocolumn,letterpaper]{article}

\usepackage{iccv}
\usepackage{times}
\usepackage{epsfig}
\usepackage{graphicx}
\usepackage{amsmath}
\usepackage{amssymb}
\usepackage[accsupp]{axessibility}  % Improves PDF readability for those with disabilities.
\usepackage{algorithm,algorithmic}
\usepackage[numbers,sort,compress]{natbib}
\usepackage[breaklinks=true,bookmarks=false]{hyperref}

\iccvfinalcopy % *** Uncomment this line for the final submission

\def\iccvPaperID{6245} % *** Enter the ICCV Paper ID here

\ificcvfinal\fi
\usepackage[usenames,dvipsnames]{xcolor}
\usepackage{url}
\usepackage{enumitem}
\makeatletter

\newcommand{\Rmnum}[1]{\expandafter\@slowromancap\romannumeral #1@}
\makeatother

\newcommand{\myparagraph}[1]{\vspace{0.1em}\noindent\textbf{#1}}

\newcommand{\shelf}{\textbf{\textit{Shelf}}\xspace}
\newcommand{\association}{\textbf{\textit{Association}}\xspace}

\begin{document}

%%%%%%%%% TITLE
\title{Shape-aware Multi-Person Pose Estimation from Multi-View Images}

\author{Zijian Dong$^{1}$ \quad Jie Song$^{1}$ \quad Xu Chen$^{1,2}$ \quad Chen Guo$^{1}$ \quad Otmar Hilliges$^{1}$ \\
 $^1$ETH Z{\"u}rich \quad 
 $^2$Max Planck Institute for Intelligent Systems, T{\"u}bingen \\
}


\maketitle
\ificcvfinal\thispagestyle{empty}\fi

%%%%%%%%% ABSTRACT
\begin{abstract}
In this paper we contribute a simple yet effective approach for estimating 3D poses of multiple people from multi-view images. Our proposed coarse-to-fine pipeline first
aggregates noisy 2D observations from multiple camera views into 3D space and then associates them into individual instances based on a confidence-aware majority voting technique. The final pose estimates are attained from a novel optimization scheme which links high-confidence multi-view 2D observations and 3D joint candidates. Moreover, a statistical parametric body model such as SMPL is leveraged as a regularizing prior for these 3D joint candidates. Specifically, both 3D poses and SMPL parameters are optimized jointly in an alternating fashion. Here the parametric models help in correcting implausible 3D pose estimates and filling in missing joint detections while updated 3D poses in turn guide obtaining better SMPL estimations. By linking 2D and 3D observations, our method is both accurate and generalizes to different data sources because it better decouples the final 3D pose from the inter-person constellation and is more robust to noisy 2D detections. We systematically evaluate our method on public datasets and achieve state-of-the-art performance. The code and video will be available on the project page: \href{https://ait.ethz.ch/projects/2021/multi-human-pose/}{\color{magenta}{https://ait.ethz.ch/projects/2021/multi-human-pose/}}.

\end{abstract}

%%%%%%%%% BODY TEXT
\newcommand{\figureTeaser}{

\begin{figure}
\begin{center}
\includegraphics[width=\linewidth]{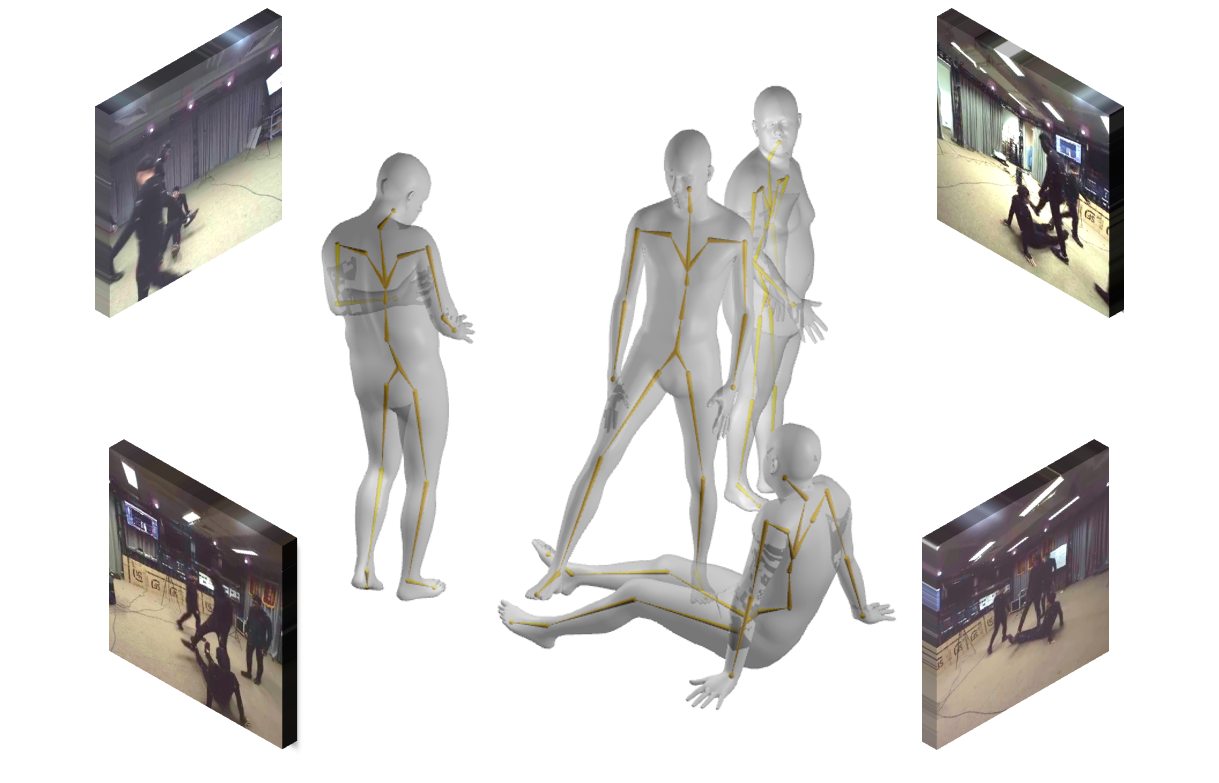}
% \fbox{\rule{0pt}{2in} \rule{.9\linewidth}{0pt}}
\end{center}
   \caption{\textbf{Shape-aware multi-person pose estimation:} We propose a novel pipeline for robust recovery of 3D poses and shapes of multiple people from a few camera views. A formulation that links 2D and 3D observations and that is regularized via a parametric body model is robust to noisy and missing 2D detections. Articulated poses can even be recovered under heavy occlusion.}

\label{fig:teaser}
\end{figure}
} 

\newcommand{\figurePipeline}{

\begin{figure*}
\begin{center}
\includegraphics[width=\linewidth]{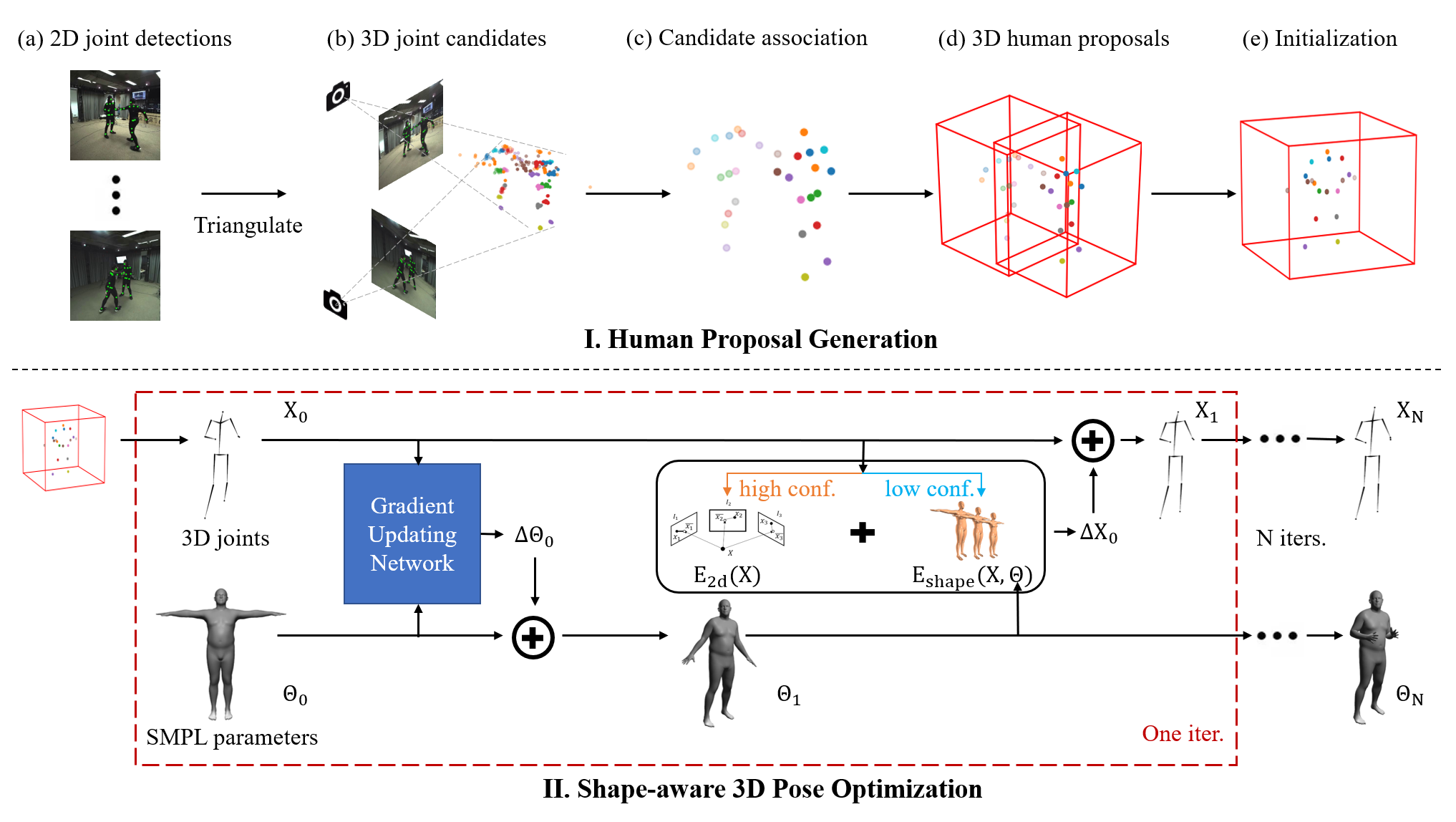}
% \fbox{\rule{0pt}{2in} \rule{.9\linewidth}{0pt}}
\end{center}
   \caption{\textbf{Pipeline structure.} Stage \Rmnum{1}: (a): We apply a 2D human pose estimation method~\cite{cao2018openpose} to obtain 2D joint candidates. (b): 2D candidate pairs with the same part label are triangulated into 3D space to produce 3D joint candidates. (c): A confidence-aware voting-based algorithm is used for clustering joint candidates from partial observations. (d): The position of human instances can be  detected based on a reliable joint. 
   (e): For each 3D human proposal, we project it back into the image space and leverage the part affinity field feature (PAF~\cite{cao2018openpose}) to filter the joint candidates from closely interacting people and obtain initial 3D pose proposals. Stage \Rmnum{2}:
   We refine the initial 3D poses $X_0$ by optimizing a 2D-3D objective. Both 3D poses $X$ and SMPL parameters $\Theta$ are optimized alternatively. For each iteration, the 3D joint locations $X$ are optimized by a 2D re-projection error when the corresponding 2D joint detections are of high confidence. To obtain kinematically plausible poses, we leverage updated SMPL estimation for regularizing the low-confidence 3D joint candidates. The SMPL parameters $\Theta$ are encouraged to align to the updated 3D poses in each iteration via a learned gradient updating network. After a small number of iteration, our method can generate complete and accurate 3D human poses and output SMPL parameters.}

   %We refine the intial 3D poses by optimizing an energy function which includes a multi-view re-projection term and a 3D body model fitting term. Both terms are optimized alternatively. After optimization, our method can generate complete and accurate 3D human poses and output SMPL mesh models~\cite{loper2015smpl}.}
\label{fig:pipeline}
\end{figure*}
}

\newcommand{\figureAblationsmpl}{

\begin{figure}
\begin{center}
\includegraphics[width=\linewidth]{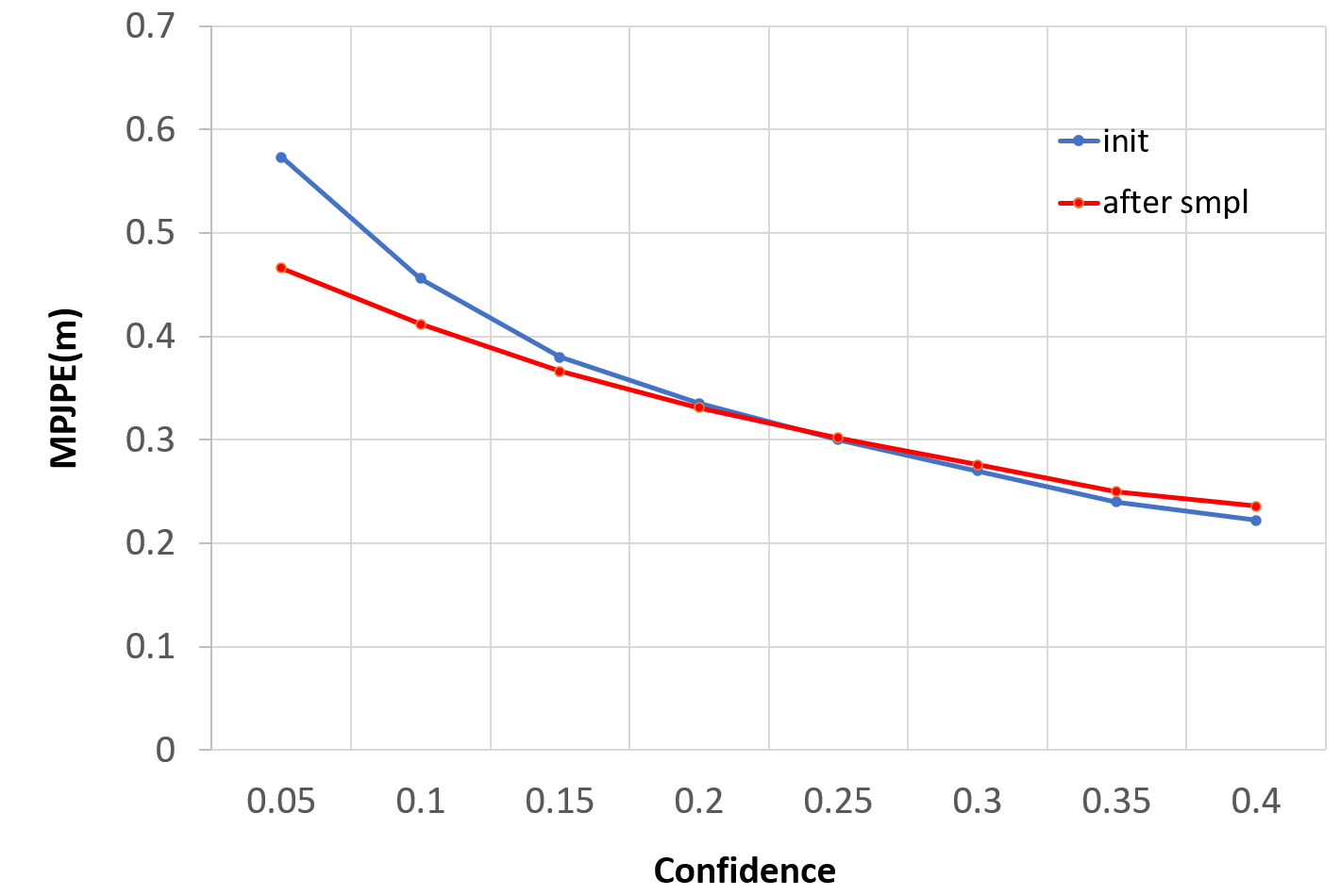}
% \fbox{\rule{0pt}{2in} \rule{.9\linewidth}{0pt}}
\end{center}
   \caption{ \textbf{Comparison of MPJPE (m) between our initial proposals (init) and the predicted poses after regularizing with SMPL (after smpl).} The horizontal axis represents the associated confidence of the 3D joints from the initial proposal.  When the confidence is under 0.25,  our algorithm achieves better performance  after  regularizing  the  3D  pose  via  a  parametric body model.}
%   For joints with high confidence(>0.25), the effect is not obvious. Thus we only optimize them with 2D reprojection errors without regularization
\label{fig:ablationsmpl}
\end{figure}
}

\newcommand{\figureResultpose}{

\begin{figure*}
\begin{center}
\includegraphics[width=\linewidth]{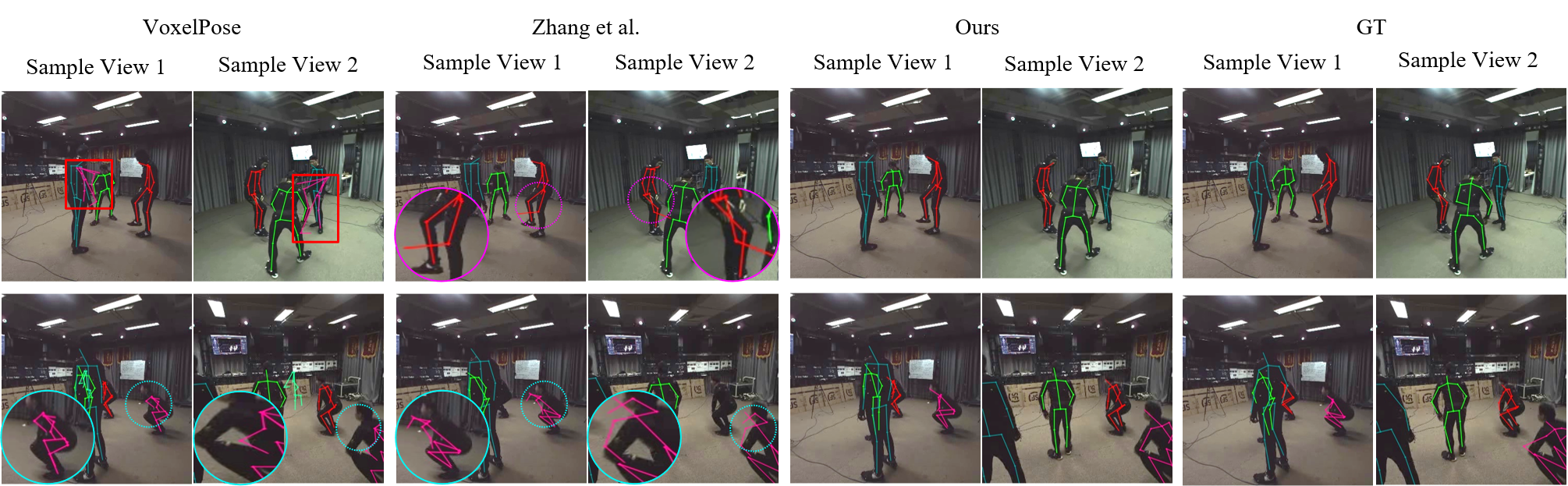}
% \fbox{\rule{0pt}{2in} \rule{.9\linewidth}{0pt}}
\end{center}
   \caption{\textbf{ Qualitative comparison with VoxelPose~\cite{tu2020voxelpose} and  Zhang et al.~\cite{zhang20204d} on the Association Dataset.} Different colors stand for different types of errors: red rectangles for extra actors; blue circles for incorrect joint positions; purple circles for abnormal human poses. Each row is an independent sample and the results are the projected 2D poses from the 3D predictions. Our method is more accurate compared to others especially in challenging scenarios with strong occlusions or highly articulated poses.}
\label{fig:resultpose}
\end{figure*}
}

%  Samples of this dataset with (a): challenging pose; (b): strong occlusion (c): high-speed motion.The predicted poses of our method is more accurate without abnormal human poses on these three challenging scenarios

\newcommand{\figureAblationOptimization}{

\begin{figure*}
\begin{center}
\includegraphics[width=\linewidth]{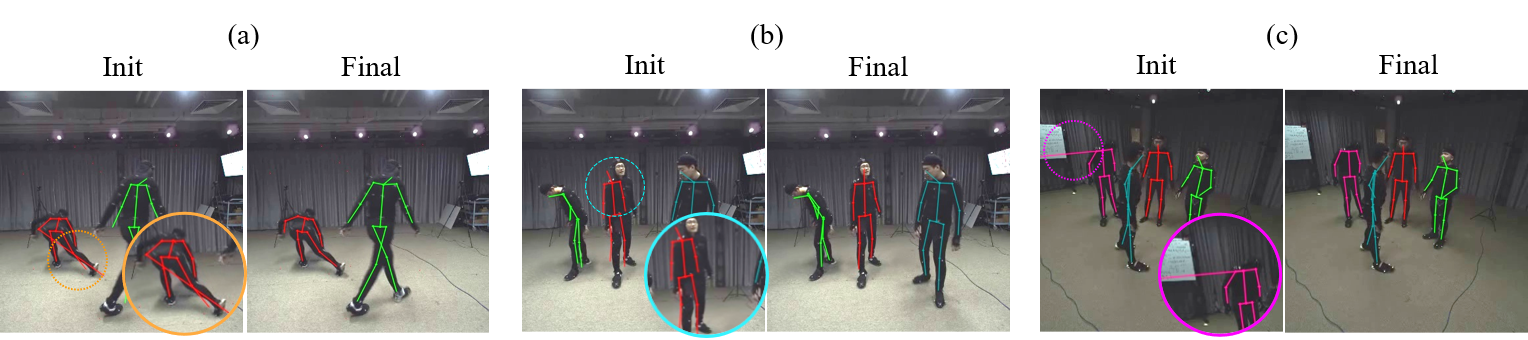}
% \fbox{\rule{0pt}{2in} \rule{.9\linewidth}{0pt}}
\end{center}
   \caption{ \textbf{Comparison between our initial poses and final poses after optimization.} (a) The orange circle is set for missing joints (note that we set the missing joints to the origin point) (b) The blue circle is set for incorrect joint predictions;(c) The purple circle is set for abnormal human poses. After the shape-aware optimization, our method generates more complete and accurate 3D human poses.}
\label{fig:ablationoptimization}
\end{figure*}
}

\newcommand{\figureDemo}{

\begin{figure*}
\begin{center}
\includegraphics[width=\linewidth]{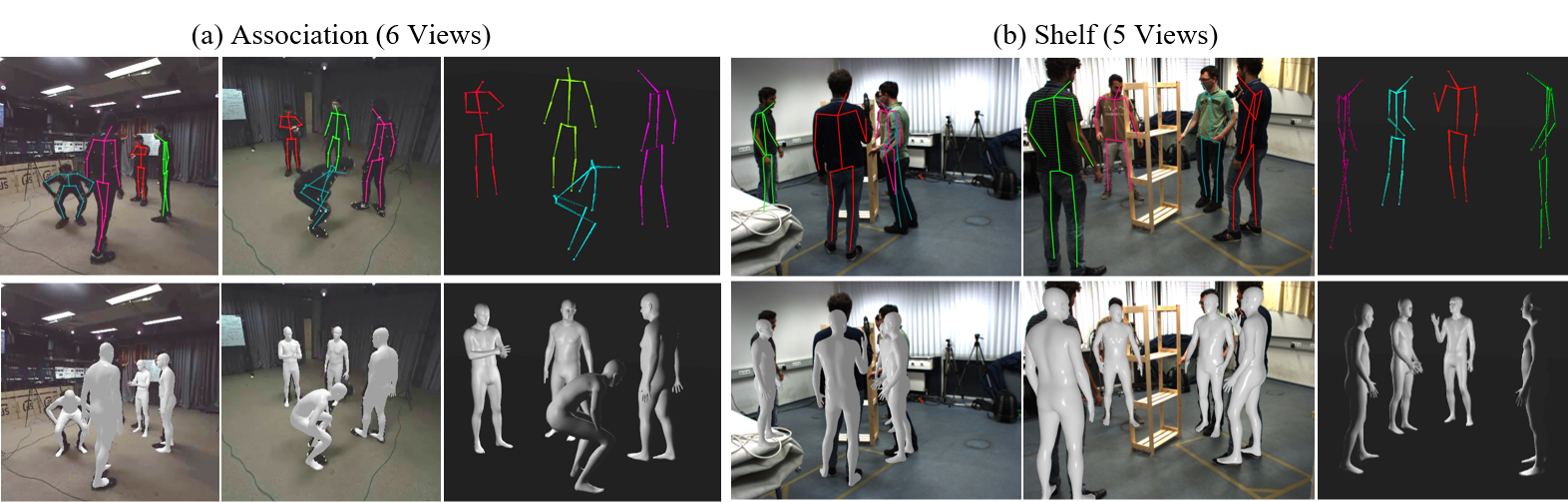}
% \fbox{\rule{0pt}{2in} \rule{.9\linewidth}{0pt}}
\end{center}
   \caption{ \textbf{ Results of our method on (a)  Association Dataset(6 views) and (b) Shelf (5 views).} The 2D reprojections of predicted 3D skeletons and SMPL mesh model are shown in two different views (left and middle); Skeletons and SMPL models of all the actors in 3D are demonstrated in the right column. More results can be found in the supplementary.}
\label{fig:demo}
\end{figure*}
}

\newcommand{\figureFitting}{

\begin{figure}
\begin{center}
\includegraphics[width=\linewidth]{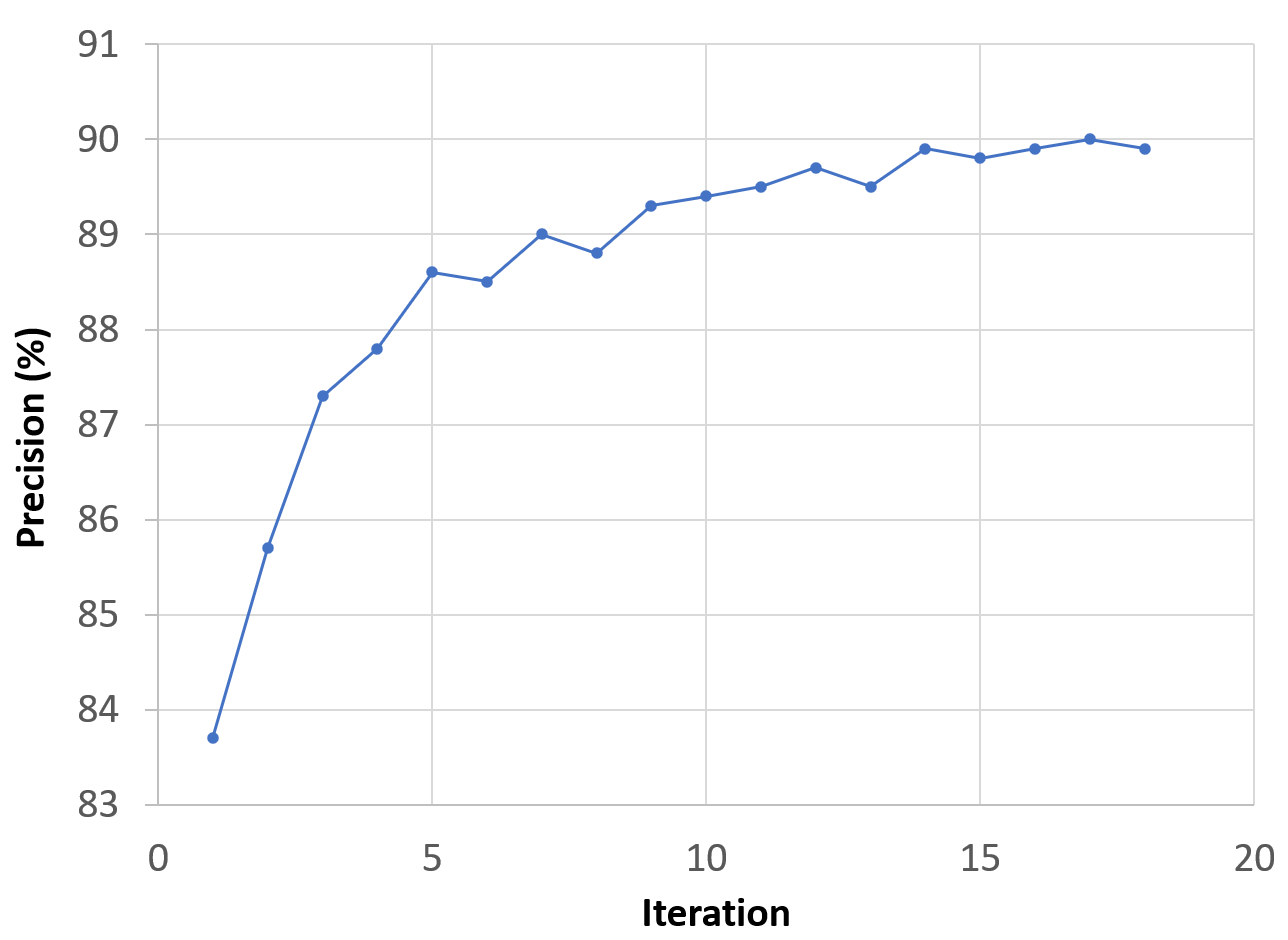}
% \fbox{\rule{0pt}{2in} \rule{.9\linewidth}{0pt}}
\end{center}
   \caption{\textbf{The precision as a function of iteration count on Association Dataset.} After 10 steps, our alternating optimization converges and finally achieves a precision of over 90 percent. }
\label{fig:fitting}
\end{figure}
}

\newcommand{\figureBbox}{

\begin{figure}
\begin{center}
\includegraphics[width=\linewidth]{figures/ablation_bbox.png}
% \fbox{\rule{0pt}{2in} \rule{.9\linewidth}{0pt}}
\end{center}
   \caption{\textbf{Study of the robustness of our human proposal generation method} We change the number of views where the hip is observed and draw the precision and recall of our method corresponding to different views in the red line and blue line respectively. Note that our method on 4D Association Dataset achieves a precision of 97.8(red point) and a recall of 94.2(blue point) and the average number of views where the hip is observed is 4.8. This proves that our method is robust when a few 2d detections are missing or noisy. }
\label{fig:bbox}
\end{figure}
}

\section{Introduction}

Markerless human motion capture is one of the fundamental problems in computer vision. In recent years much progress has been made in estimating the configuration of the human body in 2D~\cite{newell2016stacked,wei2016cpm, cao2018openpose,he2017mask,pishchulin2015deepcut} and 3D~\cite{bogo2016keep,kanazawa2018end,martinez2017simple,mehta2017vnect,zanfir2018monocular} from a single RGB image as input. 
However, if we consider settings in which multiple people are depicted and in particular if these people are interacting with each other at close range, we can expect a multitude of difficulties due to the heavy and complicated occlusions and depth ambiguities. 
To robustly estimate the poses of such groups, multi-camera setups are indispensable to provide additional observations from different views which can resolve occlusion and provide stereo cues for 3D estimation.  

% It is noteworthy though that these methods rely on first cropping the local region around the depicted person before predicting the pose.

\figureTeaser

%What others do and why it doesn't work well enough
Due to the real-world importance of this problem, several recent approaches have attempted to predict the poses of multiple people, observed from multiple cameras~\cite{chen2020multi,dong2019fast,huangend,zhang20204d,tu2020voxelpose,song2017thin}. 
Such methods can loosely be categorized into two groups. The first group formulates the problem as a cross-view matching and association problem~\cite{chen2020multi,dong2019fast,zhang20204d}.
% The problem has been formulated as a cross-view matching and association problem~\cite{chen2020multi,dong2019fast,zhang20204d}.
For example, Zhang et al.~\cite{zhang20204d} introduce an optimization formulation that attempts to jointly solve the per-view parsing and cross-view matching problems as an instance of the multicut problem. The formulation is based on an association graph that links joints within and across multiple views. 
While the formulation is elegant, in practice it requires traversing the dense, cyclic association graph which results in an NP-hard problem. To attain a computationally tractable approach, the authors revert to a greedy heuristic which is sensitive to noisy 2D joint detections and imperfect visual features which limits the accuracy of the method. 

Other methods, such as Tu et al.~\cite{tu2020voxelpose} combine the features from individual camera views into a 3D voxel space. This volume is then segmented into sub-volumes by a learned person detector. 
The final 3D human pose configurations are regressed from these sub-volumes. 
Because the pipeline can be trained end-to-end, high accuracy can be achieved if the training and test distributions are similar.  
However, owing to the reliance on the volumetric feature representation -- which encapsulates the joint distribution of different individuals, their location in 3D space, the camera setup and even the 2D joint detections -- learning such a representation requires a vast amount of data. In the absence of a large corpus of annotated multi-people, multi-view data, such methods face generalization issues and are sensitive to distribution shifts (\tabref{table:Association}).

Embracing this challenging problem, we propose a simple yet effective coarse-to-fine pipeline to estimate 3D multi-person poses from multi-view images. 
Our method combines concepts from both bottom-up and top-down methods. To avoid having to solve the association problem with partial local evidence, we aggregate initial 3D pose proposals in a 3D feature space. Our first insight is that, in pose estimation, the uncertainty associated with the 2D features (i.e., joint detections) can be trusted more than in many other computer vision domains due to semantics. Thus we forgo neural-network based classifiers and propose a simple confidence-aware majority voting technique to obtain initial 3D proposals. We experimentally show that it is more robust to distribution differences in terms of human poses, location of individuals and cameras in space and thus leads to better generalization behavior. This coarse 3D localization step is followed by a refinement step to correct poses and fill in missing joints via an optimization scheme that leverages multi-view constraints directly, where high confidence 2D observations are available, and regularizes the 3D pose via a parametric body model. 

More precisely, the first part of our pipeline consists of triangulating the 3D coordinates of all pairs of 2D detections with the same part label. This is followed by a confidence-aware majority voting technique to cluster the proposals. The technique is based on the insight that if a joint has been seen and predicted accurately (i.e., with high confidence) in several views, then there will be a dense cluster of 3D candidates for that joint and low confidence, isolated candidates can be discarded. Furthermore, we leverage the observation that certain joints, for example, the hip, are detected more reliably than end-effectors and can be used as a heuristic to decide the number and location of individual humans. While simple, experiments (\tabref{table:Bboxinstance}) show that our approach outperforms the SOTA learning-based method~\cite{tu2020voxelpose} and matching-based method~\cite{zhang20204d} in terms of the detection performance.

The second part of our pipeline refines the initial 3D estimates based on a novel 2D-3D objective (Eq.\eqref{eq:err_total}). In our formulation, we optimize the 3D joint locations directly by minimizing the 2D re-projection error if the corresponding 2D joint detections are of high confidence (Eq.\eqref{eq:err_reprojection}). 
To regularize the fitting procedure and to attain complete and kinematically plausible poses we leverage SMPL for low-confidence 3D candidates (Eq.\eqref{eq:err_regularize}). 
Importantly, the SMPL parameters are aligned directly to the updated 3D observations (current state of the 3D joint locations). For this we use the learned per-parameter gradient method (Eq.\eqref{eq:lgd}).
This approach is fundamentally different from most existing approaches~\cite{bogo2016keep} that fit SMPL parameters directly to 2D observations. We experimentally show that the triangulated 3D joints are more accurate than the PCA-based SMPL skeleton -- if they stem from confident 2D observations (\figref{fig:ablationsmpl}). 
% \todo{provide some evidence; I believe we have an experiment that shows this.} 
%
Both initial 3D pose proposal and SMPL parameters are optimized in an alternating manner (Alg. \ref{alg:inference}). This is motivated by the insight that a good estimate of 3D poses helps in fitting SMPL, while better SMPL estimates make 3D poses more robust.
% The second part of our pipeline refines the initial 3D estimates in order to correct implausible pose estimates and to fill in the location of missing joints. This refinement step is based on an energy formulation which optimizes both initial 3D pose proposal $X$ and SMPL parameters jointly (~Eq.\eqref{eq:err_total}). Different from prior methods which optimize SMPL parameters by fitting them to 2D observations via a re-projection error, we leverage multi-view constraints directly where the 2D observations are available and have high confidence and regularizing the 3D pose via a parametric body model where 2D detections are missing or have low associated confidence. The key reason to not only estimate SMPL parameters is that the triangulated 3D joints are more accurate than the PCA based SMPL skeleton if they stem from confident 2D observations (~\tabref{fig:ablationoptimization}). Both initial 3D pose proposal and SMPL paramters are optimized in an alternating manner (~Alg. \ref{alg:inference}). This is motivated by the insight that a good estimate of 3D pose proposal helps in fitting SMPL, while better SMPL estimates make 3D pose proposal more robust (~\figref{fig:fitting}).
Finally, detailed experiments are performed to demonstrate that both components improve the robustness and accuracy of the pose estimation task. In summary, our main contributions are:
% Finally, detailed experiments are performed to demonstrate that both components improve the robustness and accuracy of 3D multi-person pose estimation from multi-view images qualitatively and quantitatively.  
% In summary, our main contributions are:
\begin{itemize}
    \item A coarse-to-fine confidence-aware pipeline to aggregate noisy 2D observations from all camera views into 3D space and associate them into individual instances.
    
    \item A novel refinement pipeline which optimizes 3D poses and their corresponding SMPL models in an alternating fashion. The parametric models help in regularizing low-confidence 3D poses while updated 3D poses in turn guide the SMPL parameter estimation. 
      \item Our method is general since we only leverage off-the-shelf 2D pose detector and body pose priors distilled from motion capture datasets. SOTA performance is achieved on public datasets.

\end{itemize}

\section{Related Work}
A vast amount of work on single-person 3D pose estimation from monocular \cite{pavlakos2017harvesting,remelli2020lightweight,elhayek2015efficient,vlasic2008articulated,de2008performance,kehl2006markerless,remelli2020lightweight,corazza2010markerless} and multi-view  \cite{iskakov2019learnable,he2020epipolar, qiu2019cross} images exists. Since we study the setting of multi-person  pose estimation from multiple views \cite{liu2013markerless,joo2017panoptic,dong2019fast,zhang20204d,joo2018total,belagiannis20143d,belagiannis20153d,ershadi2018multiple}, the focus of this literature review is on multi-person pose estimation.\\ 

%  Since such methods benefit from accurate annotations of 3D pose and shape, multi-person 3D pose estimation from multiple views is becoming an important topic~\cite{liu2013markerless,joo2017panoptic,dong2019fast,zhang20204d,joo2018total,belagiannis20143d,belagiannis20153d,ershadi2018multiple}.

% More recently, the setting of multi-person 2D pose estimation from a single view has received significant attention~\cite{pishchulin2012articulated,he2017mask,fang2017rmpe,li2019crowdpose,cao2018openpose,kreiss2019pifpaf,wang2020deep}.

\myparagraph{Multi-Person 2D Pose Estimation}\\
% The recent approaches for multi-person 2D pose estimation can mainly be grouped either to the top-down or bottom-up categories. The top-down approach starts by identifying and roughly localizing individual person instances via bounding box based object detector, followed by single-person pose estimation inside the bounding box~\cite{he2017mask,fang2017rmpe,li2019crowdpose}. 
A natural approach to multi-person 2D pose estimation is to detect people first and then estimate the body pose independently. Pishchulin et al.~\cite{pishchulin2012articulated} employs a pictorial structure model to locate the person and subsequently estimate the pose. More recent top-down approaches also follow a similar strategy but instead use CNN-based person detectors and pose estimation models ~\cite{wei2016convolutional,he2017mask,fang2017rmpe,li2019crowdpose, wang2020deep}. 

% As an representative example, He et al.~\cite{he2017mask} extended the person detector by adding a branch for predicting body pose in parallel. This modification makes training the complete pipeline in an end-to-end manner possible and boosts the performance.

In contrast, bottom-up approaches~\cite{pishchulin2015deepcut,newell2016associative,papandreou2018personlab,cao2018openpose,cheng2020higherhrnet,song2019end} begin with localizing identity-free body part proposals and associates them into individual instances.
The seminal work by Pishchulin et al.~\cite{pishchulin2015deepcut} proposed a framework that jointly labels part candidates and also associates them into individual people. More recently, Cao et al.~\cite{cao2018openpose} introduced a representation of pairwise scores via the so-called Part Affinity Fields (PAF). The authors demonstrated that PAFs are able to provide effective features for the part association that a simple greedy bipartite parse can be directly applied achieving state-of-the-art results.

% Our setting is that of multi-person 3D pose estimation. Here we categorize existing work into single-view methods and multi-view methods.
\myparagraph{Multi-Person 3D Pose Estimation}\\
 When only one camera is available, the problem is under-determined since many 3D poses may correspond to the same 2D pose. Leveraging the learning-based method, 3D poses can be recovered by lifting detected 2D poses~\cite{rogez2017lcr,rogez2019lcr,zhen2020smap}, or directly regressing 3D poses~\cite{wang2020hmor,fabbri2020compressed, zanfir2018deep,benzine2020pandanet}, or by fitting parametric human body models~\cite{jiang2020coherent,zanfir2018deep}. However, the reconstruction accuracy of these methods is limited due to the depth ambiguities and strong occlusions when multiple humans are close to each other.  

Most closely related to ours are methods that leverage multi-view images. A straightforward approach for this problem is to find correspondences across views, either leveraging high-level features such as human instances, or low-level features such as joints. Early work implicitly solves this matching and parsing problem by leveraging 3D pictorial structure models, in which nodes represent 3D locations of body joints and edges encode pairwise relations between them~\cite{belagiannis20153d}. However, such methods are computationally expensive due to the large state space in 3D. Joo et al.~\cite{joo2017panoptic} rely on local features from dense multi-view images to vote for possible 3D joint positions, which can be seen as an implicit form of matching.

The method proposed by Dong et al.~\cite{dong2019fast} first performs per-view person parsing, followed by a cross-view person matching via a convex optimization method constrained by cycle consistency.  In~\cite{zhang20204d}, the authors formulate parsing, matching, and tracking in a unified graph optimization framework to simultaneously address 4D information. In contrast to these matching-based methods, a recent work~\cite{tu2020voxelpose} directly localizes all people and estimates their corresponding 3D poses in 3D voxel space. Due to the reliance on the 3D feature representation as input for subsequent learning-based steps, this method faces challenges in generalization with different configurations of people, poses and cameras. In our work, we propose a simple yet effective pipeline that triangulates joint candidates and associates them into individual instances via a simple confidence-aware voting scheme. A 2D-3D optimization technique, optimized via learned gradient descent, produces highly accurate 3D pose estimates. We show that this pipeline outperforms both matching-based approaches and end-to-end learning methods.

\section{Method}

\figurePipeline

\figref{fig:pipeline} provides an overview of our proposed approach, which contains two  stages: 3D human proposal generation and shape-aware 3D pose optimization. 
%
%
% In the first stage (\figref{fig:pipeline}, \Rmnum{1}), an off-the-shelf 2D human pose estimation algorithm \cite{cao2018openpose} is applied to images from different views (\figref{fig:pipeline}, (a)). Then all 2D detections with the same part label from pairs of views are triangulated into 3D space to produce 3D joint candidates (\figref{fig:pipeline}, (b)).
% %
% Based on these 3D candidates, a confidence-aware voting-based technique is applied to cluster joint candidates from noisy observations (\figref{fig:pipeline}, (c)) and the position of human instances is determined by considering a reliable joint (\figref{fig:pipeline}, (d)).  
% %
% To generate pose proposal for each detected human instance,  a 3D bounding box around its hip is placed and is projected back to the images. The image observations surrounding the body's limbs are used to filter joint candidates from closely interacting people (\figref{fig:pipeline}, (e)). 

In the first stage, we generate 3D joint candidates by triangulating 2D human pose estimates from different views. Then a confidence-aware voting-based technique is applied to cluster joint candidates from noisy observations and determine human instances. To generate a pose proposal for each human instance, a 3D bounding box is placed around its hip and is projected back to the images. The image observations surrounding the body's limbs are used to filter joint candidates from closely interacting people. 

% %
% In the second stage (\figref{fig:pipeline},  \Rmnum{2}), to refine the initial pose $X_{0}$, an energy formulation which includes a multi-view re-projection term $E_{2d}(X)$ and a 3D body model fitting term $E_{shape}(X,\Theta)$ is introduced. Both 3D poses $X$ and SMPL parameters $\Theta$ are  optimized  jointly  in  an  alternating  manner (Alg.~\ref{alg:inference}).
% %
% For each iteration, the gradient updating network first takes current 3D poses $X$ and SMPL estimation $\Theta$ as input to guide updating SMPL prediction. Then the current 3D poses $X$ are optimized by minimizing the multi-view re-projection error when they stem from confident 2D observations and the updated SMPL prediction are leveraged for regularizing low-confidence or missing 3D joint candidates. After a small number of iterations, our method can generate complete and accurate 3D human poses.

In the second stage, an energy formulation which includes a multi-view re-projection term $E_{2d}(X)$ and a 3D body model fitting term $E_{shape}(X,\Theta)$ is introduced, to refine the initial pose $X_{0}$. Both 3D poses $X$ and SMPL parameters $\Theta$ are  optimized  jointly  in  an  alternating  manner. For each iteration, the gradient updating network first takes current 3D poses $X$ and SMPL estimation $\Theta$ as input to guide updating SMPL prediction. Then the current 3D poses $X$ are optimized by minimizing the multi-view re-projection error when they stem from confident 2D observations and the updated SMPL prediction is leveraged for regularizing low-confidence or missing 3D joint candidates. After a small number of iterations, our method can generate complete and accurate 3D human poses.

\subsection{3D Human Proposal Generation}

\label{sec:human_proposal}
One of the main challenges for multi-person pose estimation from multiple views is to associate 2D poses from different views with consistent identities.
Prior matching-based work~\cite{zhang20204d}, is sensitive to imperfect 2D detections due to its local heuristic, and purely learning-based methods~\cite{tu2020voxelpose,huangend} are prone to overfitting. 
In contrast, we propose an effective approach to generate initial 3D pose proposals based on a confidence-aware voting technique, operating in the global 3D space of joint candidates that have been triangulated from pairs of 2D noisy detections.
\\

% The main objective of multi-view data association is to group 2d poses from different views with people's identities to connect the 2d and 3d pose estimation. This task is challenging due to several reasons. First of all, the 3d supervision is quite difficult to obtain and current dataset always has small sizes. This makes learning-based method difficult to generalize to different datasets.  Secondly, there are always errors and missing joints in the estimated 2d poses which cause wrong matching pairs. The third reason is that people are always occluded by each other and this can significantly influence the accuracy of 2d huamn detection. 

\myparagraph{3D Joint Candidates Reconstruction.}\\
To reconstruct 3D joint candidates, we first run an off-the-shelf 2D human pose detector~\cite{cao2018openpose} on each input image to generate 2D joint detections (\figref{fig:pipeline}, (a)). 
Then pairs of joints with the same label from different views are triangulated into 3D joint candidates (\figref{fig:pipeline}, (b)). 
We use standard linear algebraic triangulation~\cite{hartley2003multiple}, solving the linear system defined on the homogeneous 3D coordinate vector $\Tilde{y_j}: A_j\Tilde{y_j} = 0$, where $A_j \in \mathbb{R}^{(2C,4)}$ is a matrix composed of the components from the projection matrices and 2D poses. In our case, we perform the triangulation from each pair of 2D poses and set $C$ to 2.

%  we first generate all possible pairs of 2d poses from various views based on the detected 2d human body part candidates. For each pair of 2d poses, we use a linear algebraic triangulation approach\cite{hartley2003multiple} to extract the corresponding 3d poses. This method finds the 3d pose coordinate of a joint $y_j$ by solving  the  system
% of equations on homogeneous 3d coordinate vector of the
% joint $\Tilde{y_j}$:
% \begin{equation}
%         A_j\Tilde{y_j} = 0
% \end{equation}
% where $A_j \in \mathbb{R}^{(2C,4)}$ is a matrix composed of the components from the projection matrices and 2D poses. (see \cite{hartley2003multiple} for more details) \\

\myparagraph{Candidates Association.} \\
The next step is to associate triangulated 3D joint candidates into individual instances. Our insight for the association is simple: since we triangulate pairs of joint detections, joints that are visible in several views produce  dense clusters of 3D candidates. 
%% commented this out since I couldn\t parse it - if its important revise and bring it back
%The candidates in the cluster must belong to the same joint of the same person and their 2D observations are regarded as a match. 
Based on this observation, we propose an efficient and effective voting-based algorithm. 

For the set $C_i$ of all the 3D joint candidates with part label $i$, we initialize an empty set $S_i$ and update it iteratively. In each iteration, we first find the point $p_k$ with the highest confidence in $C_i$. Next, a subset $s_k \in C_i$ containing all the neighboring 3D candidates around $p_k$ with a distance less than threshold $\rho$ %\oh{greater than, less than?} 
is selected. We add $s_k$ to $S_i$ and remove $s_k$ from $C_i$. 
We repeat the above until $C_i$ becomes the empty set. Since outliers usually stem from falsely associated 2D detections or wrong detections in one specific view, there will only be few neighboring candidates around them. We thus eliminate clusters with less than three points. For the remaining clusters, we use their center to represent its position in 3D.

% After reconstructing the 3D joint candidates, we now need to filter out  outliers. The principle for selection is simple: if a joint has been seen and predicted accurately by several views, then there will be a dense cluster of 3D candidates for that joint. Based on this observation, we propose an efficient and effective voting based algorithm. For the set $C_i$ of all the 3D joint candidates with part label $i$, we initialize an empty set $S_i$ and update it iteratively. In each iteration, we first find the point $p_k$ with the highest confidence in $C_i$. The confidence is from the 2D detections by which the 3D joint candidate is triangulated. Next, a subset $s_k \in C_i$ containing all the neighboring 3D candidates around $p_k$ with a distance threshold $\rho$ is selected. We add $s_k$ to $S_i$ and remove $s_k$ from $C_i$. We repeat the above until $C_i$ becomes the empty set. 
% %
% Since outliers usually stem from falsely associated 2D detections, there will only be few neighboring candidates around them. We thus eliminate clusters with less than three points. For the remaining clusters, we use their center to represent its position in 3D. 
% we distinguish the correct 3d poses is that, the correct 3d poses are calculated by 2d poses belonging to the same person and therefore these 3d poses are always adjacent to each other. Therefore, if the distance between a pair of 3d poses are very small, they must belong to the same joint of the same person and their 2d poses are regarded as a match.

\myparagraph{Human Proposal Generation.}\\
The filtered 3D joint candidates need to be associated with individual instances. We experimentally find that the hip joint is one of the most reliable parts and can be leveraged to robustly decide the number of instances and also the location of each instance. Hence, we simply place a 3D bounding box with fixed size and orientation using the hip candidates as anchors. Furthermore, we keep the anchors whose corresponding 3D bounding box contains more than 90\% of body parts and whose average confidences are larger than an empirically derived threshold. These bounding boxes may still contain joints of other closely interacting people. In order to distinguish them, we project the 3D bounding box back to the image space and use the part affinity fields~\cite{cao2018openpose} to determine which 3D joints belong to other person instances.

% it's easy to adapt the network in ~\cite{tu2020voxelpose} for localizing all people. However, we find this learning-based approach easily overfitted to the training dataset and can't generalize to people with unseen 3d poses. Thus we propose a simple but effective pipeline for detecting cuboid proposal. We represent a cuboid proposal by a 3d bounding box with fixed size and orientation. This is reasonable because the size of people in 3D space have limited variation. Based on this simple assumption, we select the locations of hips from previous grouping algorithm as the anchors, since they are easier to detect and lies in the center of humans. Furthermore, we keep the anchors whose corresponding 3D bounding box contains
% more than 90\% types of joints and whose average confidences are larger than a threshold.
% Non-Maximum Suppression(NMS) is also proposed when the distance between two anchors are quite small. We will numerically compare our 3d Human proposal with other methods in the experiment section.

\subsection{Shape-aware 3D Pose Optimization}

\label{section:optimization}
% In the previous section, we propose a robust and efficient 3d human proposal detection algorithm. But it is still too coarse with fixed size, which always contain some joints of other people when people interact with each other closely. To infer the 3d joints in the bounding box and distinguish joints from different people, we adopt a similar method based on PAF\cite{cao2018openpose} to parse humans in 2d images and project each 3d bounding box back to 2d images to select the corresponding joints from the same person in different views.

% To infer the 3d positions of the joints from their corresponding 2d estimates, a naive linear triangulation approach\cite{hartley2003multiple} similar to  \ref{section:triangulation} can be directly used. This naive algorithm assumes that the joint coordinates from each view are independent of each other and thus all make comparable contributions to the triangulation\cite{iskakov2019learnable}. However, for the multi-human pose estimation, on some views the 2d location of the joints can't be estimated reliably due to the joint occlusions and the errors in 2d pose estimation, which finally leads to unnecessary errors of the final triangulation. Furthermore, when people are occluded by each other, the 2d pose detection always fails for missing some joints or connecting parts from different people, which can finally lead to missing joints or abnormal human poses.

The initial pose proposals do not yet adhere to kinematic constraints and may have missing joints due to imperfect 2D joint detections. 
We refine these initial poses $X$ via both multi-view re-projection evidence $E_{2d}(X)$ and a parametric body model prior $E_{shape}(X,\Theta)$. The 3D poses $X$ and SMPL parameters $\Theta$ are optimized alternatively.
The re-projection term aligns 3D joints $X$ with the 2D observations for high-confidence joint detections. Whereas missing or low-confidence joints are determined by leveraging the updated SMPL estimation to regularize the 3D pose, leading to complete and kinematically plausible 3D pose estimates.
For SMPL parameters $\Theta$, they are optimized to align to the current 3D joints $X$ via a learned gradient updating network. This refinement finally leads to complete and kinematically plausible 3D pose estimates after a small number of iterations.
The alternating process is shown in \figref{fig:pipeline},  \Rmnum{2}.

% I

% To solve above mentioned problems, we propose an optimization-based method which optimizes the full 3d body poses by embedding a parametric human model. The 3d positions of joints from linear triangulation are used as the initialization of the optimization. 

% \subsubsection{Problem Definition}

% Given the initial 3D pose proposal $X$ and its 2D observation $x_{ij}$ f othe same human from all the views with indices $j\in (1\ldots K)$ , we hope to reconstruct the full 3D mesh of human bodies and accurate 3d joints with indices $i\in(1\ldots N)$ together. The 3D human mesh is encoded by the statistical SMPL body model\cite{loper2015smpl}, which is differentiable function that outputs a triangulated mesh $M(\theta,\beta)$ that takes the pose parameters $\theta \in \mathbb{R}^{23\times3}$ and the shape parameters $\beta \in \mathbb{R}^{10}$. The 3D body joints can be obtained conveniently as a linear combination of the mesh vertices and a linear regressor $W$ is usually pretrained to map the vertices to $N$ joints of interest, defined as $\bar{X}(\theta,\beta) = WM(\theta,\beta)$. Our aim is to optimize predicted 3d poses $X(X_1\ldots X_N)$ and SMPL parameters $\theta, \beta$ jointly.

% \subsubsection{Energy Function}

\myparagraph{Objectives.}\\
% if othe same human from all the views with indices $j\in (1\ldots K)$ , we hope to reconstruct the full 3D mesh of human bodies and accurate 3d joints with indices $i\in(1\ldots N)$ together.
% The 3D human mesh is encoded by the statistical SMPL body model\cite{loper2015smpl}, which is differentiable function that outputs a triangulated mesh $M(\theta,\beta)$ that takes the pose parameters $\theta \in \mathbb{R}^{23\times3}$ and the shape parameters $\beta \in \mathbb{R}^{10}$. The 3D body joints can be obtained conveniently as a linear combination of the mesh vertices and a linear regressor $W$ is usually pretrained to map the vertices to $N$ joints of interest, defined as $\bar{X}(\theta,\beta) = WM(\theta,\beta)$. Our aim is to optimize predicted 3d poses $X(X_1\ldots X_N)$ and SMPL parameters $\theta, \beta$ jointly.
% To deal with the problem of unreliable joints and incorrect joints, we drop the outliers and define the weighted reprojection error as follows:
% as a full-body constraint and it is defined as the average distance between the predicted 3d poses $X(X_1 \ldots X_N)$ and the joints $\bar{X}(\bar{X}_1 \ldots \bar{X}_N)$ from SMPL as follows:
Given an initial 3D pose proposal $X$ and its corresponding 2D observations $x_{ij}$, where $i\in(1\ldots N)$ stands for the joint label and $j\in (1\ldots K)$ represents the view indices,
we want to refine the 3D pose by leveraging multi-view constraints  where the 2D observations have high confidence:
\begin{equation}\label{eq:err_reprojection}
    E_{2D}(X) = \sum_{i=1}^N\sum_{j=1}^K w_{ij} \delta_{ij}d_{2D}(\Pi_jX_i, x_{ij}) \quad
\end{equation}
 Here $E_{2D}(X)$ denotes the re-projection error between the 2D joint projections into each view and the detected 2D joints.  
 $\Pi_j$ is the projection matrix of view $j$.
 $w_{ij}$ is the confidence of detected joint $i$ in the view $j$ and $\delta_{ij}$  is an indicator function denoting if  joint $i$ in view $j$ is discarded:
\begin{equation}
        \delta_{ij} = 
    \begin{cases}1, &  d_{2D}(\Pi_jX_i, x_{ij})<\rho_{2D} \cr  0, &\text{otherwise}
    \end{cases}
\end{equation}
where $\rho_{2D}$ is a threshold for selecting inliers. 

% \oh{consider switching notation to a more common indicator such as $\mathbb{I}_{ij}$}
To complement $E_{2D}$ we leverage a body shape term $E_{shape}$ to regularize the 3D pose via a parametric body model when 2D detections are missing or have low associated confidence. For this purpose, we use the SMPL model\cite{loper2015smpl}. It is a differentiable function that outputs a triangulated mesh $M(\theta,\beta)$ taking the pose parameters $\theta \in \mathbb{R}^{23\times3}$ and the shape parameters $\beta \in \mathbb{R}^{10}$ as input. The 3D body joints can be obtained by a linear regressor $W$ taking the mesh as input ($\bar{X}(\theta,\beta) = W(M(\theta,\beta))$).  We jointly  optimize predicted 3D poses $X$ and SMPL parameters $\theta, \beta$:
\begin{equation}\label{eq:err_regularize}
    E_{shape}(X,\theta, \beta) = \sum_{i=1}^{N} \delta(w_i) d_{3D}(X_i, \bar{X}_i(\theta,\beta))
\end{equation}
where $w_i$ is the average confidence of detected 2D joint $i$ across views. $\delta(w_i)$ is an indicator function denoting whether the initial 3D joints $i$ has high enough confidence:

\begin{equation}
        \delta(w_i) = 
    \begin{cases}1, &  w_i<\rho_{3D} \cr  0, &otherwise
    \end{cases}
\end{equation}

The final energy is the weighted sum of Eq. \eqref{eq:err_reprojection} and \eqref{eq:err_regularize}:
\begin{equation}\label{eq:err_total}
    E(X,\theta,\beta) = w_{2D}E_{2D}(X) + w_{shape}E_{shape}(X,\theta, \beta).
\end{equation}

\begin{algorithm}
\caption{- Alternative Optimization}
\label{alg:inference}
\begin{algorithmic}

    \STATE $X_0 \leftarrow $ initial 3D pose proposal \\
    \STATE $\Theta_0 \leftarrow \{\theta_0,\beta_0\} \leftarrow 0$ 
    % \STATE $\lambda \leftarrow $ learning rate
    % \STATE $\hat{x}_0 \leftarrow s_0\Pi(R_0X_0(\theta_0,\beta_0)) + t_0$ \\
    % \STATE $\mathcal{L}(\Theta_0) \leftarrow L_{reproj}(\hat{x}_0, x_{gt})$ \\
    % \STATE $\Delta \Theta_0 \leftarrow \mathcal{N}_w(\frac{\partial \mathcal{L}(\Theta_0) }{\partial \Theta_0}, \Theta_0, x_{gt})$\\
    % \STATE $\Theta_1 \leftarrow \Theta_0 +\Delta \Theta_0$
      \FOR{$n = 0,...,N-1$ }
            
      \STATE $\Theta^{(0)} \leftarrow \Theta_{n}  $   
          \FOR{$m = 0,...,M-1$}
           
          \STATE $\mathcal{L}_{shape}(\Theta^{(m)}) \leftarrow E_{shape}(X_n,\Theta^{(m)})$\\
          \STATE $\Delta \Theta^{(m)} \leftarrow \mathcal{N}_w(\frac{\partial \mathcal{L}_{shape}(\Theta^{(m)} }{\partial \Theta^{(m)}}, \Theta^{(m)}, {X_n})$
          \\
          \STATE $\Theta^{(m+1)} \leftarrow \Theta^{(m)} +\Delta \Theta^{(m)}$
        %   \STATE $L_{reproj} \leftarrow ||x_{gt} - \hat{x}_n||_1$
        
          \ENDFOR
    \STATE $\Theta_{n+1} \leftarrow \Theta^{(M)}$
    
    \STATE  $\mathcal{L}(X_n) \leftarrow w_{2d} E_{2d}(X_n) + w_{shape}E_{shape}(X_n,\Theta_{n+1})$ \\
    \STATE  $X_{n+1} = X_{n} + \lambda \frac{\partial \mathcal{L}(X_n)}{\partial X_n} $\\
      \ENDFOR
\end{algorithmic}
\end{algorithm}

% \subsubsection{Inference Pipeline}
\myparagraph{Alternating Optimization.}\\
% In $E(X, \theta, \beta)$, if $\theta,\beta$ are fixed, we can directly optimize $X$ by the gradient descent method. If $X$ is fixed, optimizing  $\theta,\beta$ is equal to fitting human shape model with 3d joints. However, traditional fitting approaches\cite{bogo2016keep} are very slow and not robust.
To optimize Eq.~\eqref{eq:err_total} we employ a custom gradient descent strategy. We start by fixing $X$ to its initial value and optimize $\theta,\beta$. 
Since fitting SMPL parameters to 3D observations is a non-convex and highly non-linear problem, this can be slow and error-prone with traditional methods such as that by Bogo et al.~\cite{bogo2016keep}.
We take inspiration from a recent 2D-3D lifting approach~\cite{song2020human} that solves the fitting of 3D human body by leveraging a neural network to predict the parameter update rule. 
We adopt a similar concept for fitting the human model to the 3D candidates. To accelerate fitting of  $\Theta = \{\theta, \beta\}$ we replace the standard gradient descent rule by a learned per-parameter update: 
\begin{equation}\label{eq:lgd}
    \Theta^{(m+1)} = \Theta^{(m)} + \mathcal{N}_w(\frac{\partial E_{shape}}{\partial \Theta^{(m)}}, \Theta^{(m)}, X) \quad 
\end{equation}
where $\mathcal{N}_w$ is a deep network parameterized by a set of weights $w$, $\frac{\partial E_{shape}}{\partial \Theta^{(m)}}$ is the partial derivative wrt $\Theta$ and $X$ is the fitting target. $\mathcal{N}_w$ is trained with samples of poses and shapes from AMASS~\cite{mahmood2019amass}. Please refer to ~\cite{song2020human} and supplementary for more details of the training process.

Once $\Theta$ is optimized, we keep it fixed and optimize $X$ via standard gradient descent. We optimize $X,\Theta$ in an alternating fashion until convergence. The overall routine is detailed as pseudo-code in Alg. \ref{alg:inference}, where $n$ is the iteration index for the overall optimization routine while $m$ is for the body fitting process.

\newcommand{\tableShelf}{
\begin{table}[h]
\centering
\footnotesize
\resizebox{\linewidth}{!}{
\begin{tabular}{lccccc}
\hline
Method        &Anno & A1 & A2 & A3 & Avg \\ \hline
VoxelPose \cite{tu2020voxelpose}                           &Yes  &99.3 &94.1 &97.6    & 97.0       \\

Huang \emph{et al.}~\cite{huangend}   &Yes   &98.8 &96.2 &97.2  & 97.3       \\

\hline

Belagiannis \emph{et al.}~\cite{belagiannis20143d} &No &66.1 &65.0 &83.2 & 71.4 \\

Belagiannis \emph{et al.}~\cite{belagiannis20153d}  &No &75.3 &69.7 &87.6 & 77.5 \\

Ershadi-Nasab \emph{et al.}~\cite{ershadi2018multiple}  &No &93.3 &75.9 &94.8 & 88.0 \\

*Zhang \emph{et al.}~\cite{zhang20204d} &No & 96.5& 86.8 &97.0&93.4\\

Dong \emph{et al.}~\cite{dong2019fast} &No & 98.6 & 93.7 &97.8 &96.7(96.9)\\
\hline

Ours(final)              & No          & \textbf{99.1} & 93.5 &\textbf{98.1} &\textbf{96.9}                              \\ \hline
\end{tabular}
}
\caption{\textbf{Evaluation on Shelf dataset.} Quantitative comparison on the Shelf dataset using a percentage of correct parts (PCP) metric. '*' denotes that the method discards temporal information from its original setting. ‘A1’-‘A3’ correspond to the results of three actors, respectively. The averaged result is in column ‘Avg’. The column 'Anno' indicates if the method relies on 3D supervision.   }
\label{table:Shelf}
\end{table}
}

\newcommand{\tableAssociation}{
\begin{table}[h]
\centering
\footnotesize
\resizebox{\linewidth}{!}{
\begin{tabular}{lccc}
\hline
Method        & Precision(\%) & Recall(\%) & F1-score(\%)  \\ \hline

VoxelPose \cite{tu2020voxelpose}      &55.1 &66.5 &60.3         \\

*Zhang \emph{et al.} \cite{zhang20204d}  & 97.1& 48.8 &65.0\\

\dag VoxelPose \cite{tu2020voxelpose}      &68.8 &79.2 & 73.6    \\
Dong \emph{et al.} \cite{dong2019fast}                           &71.0 &80.2 &75.3         \\

\hline
Ours(init)    & 83.7 & 82.8   & 83.4 \\
Ours(final)   & 90.1 & \textbf{89.0} & \textbf{89.2}                              \\ \hline
\end{tabular}
}
\caption{\textbf{Evaluation on the Association Dataset.} '*' denotes that the method discards temporal information from its original setting. '\dag' means the method uses 3D bounding box ground truth. }
\label{table:Association}
\end{table}
}

\newcommand{\tableDetection}{
\begin{table}[h]
\centering
\begin{tabular}{lccc}
\hline
Method        & Precision(\%) & Recall(\%) & F1-score(\%)  \\ \hline

VoxelPose \cite{tu2020voxelpose}      &55.1 &66.5 &60.3         \\

Zhang \emph{et al.} \cite{zhang20204d}  & 98.8& 49.7 &66.1\\
\hline

Ours(init)  & 83.7& 82.0 &\textbf{82.8}\\
\hline
\end{tabular}
\caption{\textbf{Human detection results on the Association Dataset.}  }
\label{table:detection}
\end{table}

}

\newcommand{\tableAblation}{
\begin{table}[h]
\centering
\begin{tabular}{lccc}
\hline
Method        & Precision(\%) & Recall(\%) & F1-score(\%)  \\ \hline

Ours(init)      &83.7  &82.8  &83.4 \\
Ours(final)  & \textbf{90.1} & \textbf{89.0}  &\textbf{89.2}\\
\hline

\end{tabular}
\caption{\textbf{Comparison between our initial pose proposals and the final predictions (after optimization).} Shape-aware refinement can boost both precision and recall.}
\label{table:ablation}
\end{table}

}

\newcommand{\tableComparewithTrad}{
\begin{table}[h]
\centering
\footnotesize
\resizebox{\linewidth}{!}{
\begin{tabular}{lccc}
\hline
Method        & Precision(\%) & Recall(\%) & F1-score(\%)  \\ \hline

Multi-view SMPLify~\cite{bogo2016keep} &78.3 & 77.4 & 77.8\\
% Ours(head) &81.1 &85.2 &83.1\\
Ours      &\textbf{90.1} & \textbf{89.0}  &\textbf{89.2} \\
\hline

\end{tabular}
}
\caption{\textbf{Comparison between our method and the traditional method~\cite{bogo2016keep}} Our method outperforms the SMPL-only baseline for both precision and recall.}
\label{table:ComparewithTrad}
\end{table}

}

\newcommand{\tableBboxinstance}{
\begin{table}[h]
\centering
\footnotesize
\resizebox{\linewidth}{!}{
\begin{tabular}{lccc}
\hline
Method        & Precision(\%) & Recall(\%) & F1-score(\%)  \\ \hline

VoxelPose \cite{tu2020voxelpose}  & 68.8 & 77.3  &72.8\\
Zhang \emph{et al.} \cite{zhang20204d} &99.6 &51.2 &67.6\\

% Ours(head) &81.1 &85.2 &83.1\\
Ours      &98.8  &\textbf{94.2}  &\textbf{96.4} \\

\hline

\end{tabular}
}
\caption{\textbf{Evaluation of human proposal generation on the Association Dataset.} A human proposal is valid if the error of its hip joint is less than 0.2m. Ours achieves better performance compared to other SOTA ones especially for recall.}
\label{table:Bboxinstance}
\end{table}

}

\section{Experiments}
 
\subsection{Test Datasets}
\label{section:testdataset}
We conduct experiments on two standard datasets for multi-view multi-person 3D human pose estimation, which consist of challenging scenarios including interactions between individuals with heavy occlusions.

\myparagraph{Shelf}~\cite{belagiannis20143d}
contains 3200 frames from 5 cameras. In terms of evaluation setting and evaluation metric, we follow prior work~\cite{belagiannis20143d,tu2020voxelpose,zhang20204d} and test on a separate test set and report the percentage of correctly estimated parts (PCP@0.5) to measure performance.

% we use the percentage of correctly estimated parts (PCP@0.5) to measure the performance, which is commonly used in prior work~

\myparagraph{Association Dataset}~\cite{zhang20204d} is one of the most challenging public datasets for this task. It contains 3 sequences with 2-4 closely interacting people observed from 6 cameras.
%The actors all wear black marker-suit for ground truth skeletal motion capture.
%For testing the robustness and generalization ability, all the frames are selected for testing.
Following~\cite{zhang20204d}, we use all the sequences for test and report the precision, recall and F1-score as evaluation metrics. A joint is considered correct if its Euclidean distance to the ground truth annotation is less than 0.2m. 

\subsection{Training Data Comparison}
For training, our method \textit{only} uses AMASS, a collection of 3D human bodies with varying poses ~\cite{mahmood2019amass}. For clarity, we would like to emphasize that our primary goal is to boost the robustness and effectiveness to new, entirely unseen humans and poses. Note that, learning-based methods~\cite{tu2020voxelpose,huangend} generally require direct 3D supervision via annotated pairs of multi-people, multi-view data. 
%and seriously overfits to small training dataset of this task. 
% Thus, our method can only be compared directly to methods that do not require the strongest form of 3D supervision. The performance of learning-based methods is only listed for completeness.

\figureAblationOptimization

\subsection{Ablation Study}
To validate the effectiveness of our method, we conduct a detailed analysis on both  the initial 3D pose proposal and the shape-aware pose refinement. All the experiments are conducted on the \association dataset. When comparing with the  learning-based method~\cite{tu2020voxelpose}, we deploy its model trained on the CMU Panoptic Dataset~\cite{joo2015panoptic}, which is the largest training dataset for this task.

\subsubsection{Evaluation of 3D Human Proposal Generation}
To evaluate the effectiveness of our human proposal generation technique, we compare our method with two SOTA methods~\cite{tu2020voxelpose,zhang20204d}. Following the metric of~\cite{zhang20204d}, one human proposal is valid if the error of its hip joint is less than 0.2m. As shown in~\tabref{table:Bboxinstance}, our method achieves significantly better performance compared to~\cite{zhang20204d, tu2020voxelpose}. The learning-based method~\cite{tu2020voxelpose} faces generalization issues to unseen human poses and motion. The bottom-up method~\cite{zhang20204d} suffers from low recall since the greedy algorithm is sensitive to noisy 2D joint detections. More ablation study on the proposal generation can be found in the supplementary.
\tableBboxinstance

\subsubsection{Evaluation of Shape-aware Pose Optimization}
In this section, we validate the effectiveness of our shape-aware optimization.
As seen in the last two rows in \tabref{table:Association}, the initial pose proposal can be dramatically improved under all metrics by leveraging multi-view constraints and regularization with the parametric body model.
There are two main reasons for this improvement. First, the full-body constraint of SMPL can infill missing joints and corrects implausible poses (see orange and purple circles in \figref{fig:ablationoptimization}). Furthermore, the weighted re-projection error contributes to refine the joint predictions, shown in blue circles in \figref{fig:ablationoptimization}. %We further explain these two points in the following experiments. 
%  \tableAblation

\myparagraph{Human Body Constraint.}
Note that the SMPL model is used as a full-body constraint to help infill missing joints and also to correct non-kinematic poses. We conduct the experiment to fit SMPL model to the initial 3D proposal and draw the MPJPEs of the joints before and after fitting in \figref{fig:ablationsmpl}. For joints with associated low average confidence,  regularizing the 3D pose via a parametric body model can boost performance. For joints with higher confidence (confidence$>$0.25), the effect of full-body constraint is not obvious. 
This is caused by a slight difference in the skeletal configurations between~\cite{cao2018openpose} and \cite{loper2015smpl}, which causes some systematic error. Therefore we only use the multi-view constraints to optimize high-confidence joints as described in the method section. 

% This misalignment can be further solved by using a more accurate human model which aligned more accurately with the 2d estimation.

\figureAblationsmpl

\myparagraph{Comparison with SMPL-only.} 
We compare our method with approaches that optimize only SMPL parameters to align the joint re-projections with 2D observations~\cite{bogo2016keep, huang2017towards}. 
For a fair comparison, we apply a multi-view variant of SMPLify. 
The comparison on the \association dataset is summarized in \tabref{table:ComparewithTrad}. Our method outperforms the SMPL-only baseline by a significant margin. This is in part due to the inherent approximation error stemming from  differences in the skeletal configurations between~\cite{cao2018openpose} and \cite{loper2015smpl} in part due to fitting errors. 
This experimental evidence is an important motivation for the design of our 3D pose fitting formulation. 
Furthermore, leveraging learned gradient descent for optimization leads to significant performance increases both in terms of runtime (20x speed-up, 0.1s vs 2s), convergence rate (14 vs 100 iters) and precision (90.1\% vs 78.3\%). 
\tableComparewithTrad
% \myparagraph{Alternating Optimization.}
% We optimize our energy function in alternating fashion, since more accurate and complete 3D poses can contribute to generating a more realistic human model and this human model contributes in turn to the refinement of incorrect joints or infilling of missing joints. We plot the process of our alternating optimization scheme in \figref{fig:fitting}. 
% The algorithm gradually improves the 3D poses and typically converges after 10 iterations. 

% \subsubsection{Evaluation of optimization method}

% For the effect of optimization method, the result is shown in Table \ref{table:ablation}. The optimization process is shown in Fig. \ref{fig:fitting}. After 13 epochs, our algorithm increases the precision from 83.7 of our initialization to 90.1 of our final result. Comparing with the linear triangulation, our method outperforms our initialization under both metrics. To further demonstrate the reason, we compare them qualitatively in Fig .\ref{fig:ablationoptimization}. There are two reasons for this improvement. First, the full-body constraint of SMPL contributes to deducing missing joints and correcting the abnormal human poses as shown in the orange and purple circles in Fig.\ref{fig:ablationoptimization}. Furthermore, leveraging weighted reprojection error and rejecting ourliers for optimization contributes to balancing the weight between unreliable and trustworthy 2d detections and solves the problem mentioned in the blue circle in Fig. \ref{fig:ablationoptimization}. 

% \figureFitting
\figureResultpose
\tableShelf

\tableAssociation
\figureDemo
\subsection{Quantitative Results}
\label{section:qualitativecompare}

We compare our method with SOTA methods quantitatively on the \shelf and \association datasets. The comparison on the \shelf is shown in \tabref{table:Shelf}. We achieve slightly better results compared to  methods~\cite{belagiannis20143d,belagiannis20153d,dong2019fast,ershadi2018multiple,zhang20204d} which do not rely on 3D supervision and achieve comparable performance compared to learning-based methods~\cite{tu2020voxelpose,huangend} which train the model based on this dataset. Since the test frames lack pose variations compared to the training set, this dataset is considered less challenging than the \association one, which also has a more strict evaluation metric.

% Note that for the method\cite{zhang20204d}, we only test their method without using temporal information for a fair comparison. Some methods\cite{dong2019fast,tu2020voxelpose,huangend} achieve similar results comparing to ours, 

% Since the evaluation metric(PCP@0.5) is loose and the test frames are less challenging and lack of pose variations comparing to the training dataset. We further test the generalization ability and robustness of our algorithm on a more challenging dataset(4d Association Dataset) with a more strict evaluation metric and compare with the state-of-art methods.
%  \figref{fig:resultpose} shows some representative results of the proposed approach on this dataset.

The quantitative result on the \association dataset is shown in \tabref{table:Association}. Since this is a pure test set, for SOTA learning-based method~\cite{tu2020voxelpose}, we directly deploy their trained model for test. As a result, our method outperforms this learning-based method largely, even when they use ground truth 3D bounding boxes. This shows that the learning-based method is prone to overfit to the training distribution. We also compare our algorithm with matching-based methods~\cite{dong2019fast,zhang20204d}.
Note that to have a fair comparison with~\cite{zhang20204d}, we compare with its static version only relying on images while it is originally a tracking method which leverages video information. 
We can see that these bottom-up based methods have relatively low recall due to the fact that they solve the global optimization with a greedy algorithm which is sensitive to missing 2D joint detections. 

% We also list the performance of \cite{zhang20204d} with tracking information and it largely improves the their human instance detection performance, but our method still achieve comparable results without using temporal information. 

% state-of-the-art learning-based method  \cite{tu2020voxelpose}, we directly use the same 2D pose estimator trained on the COCO dataset of our algorithm and only train the other part based on the CMU Panoptic Dataset\cite{joo2015panoptic}. As a result, our method outperforms this state-of-the-art learning-based method largely, even when they use ground truth 3D bounding box. This shows that the learning-based method overfits to the training dataset largely and can't generalize well to other datasets with pose variations.

\subsection{Qualitative Comparison}
\vspace{-0.1cm}
We show qualitative comparison in~\figref{fig:resultpose}.
Generally, our method is more accurate and robust compared to others especially in challenging scenarios with strong occlusions or when highly articulated poses are presented. Specifically, other methods tend to generate extra actors (red rectangles), abnormal (purple circles) or incorrect (blue circles) human poses.  The reason is that it is difficult for the learning-based method~\cite{tu2020voxelpose} to generalize to  unseen poses and motions. For the image-based version of~\cite{zhang20204d},  without temporal information, solving the 3D association graph is sensitive to noisy 2D joint detections. 
In~\figref{fig:demo}, we demonstrate more results of our method on both the \association and \shelf datasets. The first row is the predicted 3D poses and their projections on two views. The second row shows the predicted SMPL models in 3D space and also their projections in 2D images.

% Using information from sparse views, our method enables not only robust multi-person 3D skeleton prediction but also corresponding SMPL body models under heavy occlusions and challenging poses. Note that for all the multi-view multi-person datasets, there are no 3D human body mesh ground truths. We only evaluate the 3D body shape qualitatively by projecting them on multi-view images in Fig \ref{fig:demo}.

% To further demonstrate the advantages of our pipeline, we perform qualitative comparison with the SOTA  learning-based method~\cite{tu2020voxelpose} and bottom-up method~\cite{zhang20204d} for 3D pose estimation on challenging 4D Association Dataset.

\section{Conclusion}
In this paper, we propose an effective coarse-to-fine pipeline to estimate 3D multi-person poses from multi-view images. To avoid having to solve the association problem with local evidence, we aggregate initial 3D pose proposals in a 3D feature space and associate them into individual instances.
%which is robust to distribution differences in terms of human poses, location of individuals and camera setups in space.
This coarse 3D localization step is followed by a refinement step that corrects poses and fills in missing joints via an optimization routine leveraging multi-view constraints directly where high confidence 2D observations are available and regularizing the 3D pose via a parametric body model. We systematically evaluate our method on public datasets and SOTA performance is achieved.

{\small
\noindent\textbf{Acknowledgements:} Xu Chen was supported by the Max Planck ETH Center for Learning Systems. 
}

{\small
\bibliographystyle{ieee_fullname}
\bibliography{egbib}

\begin{thebibliography}{10}\itemsep=-1pt

\bibitem{belagiannis20143d}
Vasileios Belagiannis, Sikandar Amin, Mykhaylo Andriluka, Bernt Schiele, Nassir
  Navab, and Slobodan Ilic.
\newblock 3d pictorial structures for multiple human pose estimation.
\newblock In {\em Proceedings of the IEEE Conference on Computer Vision and
  Pattern Recognition}, pages 1669--1676, 2014.

\bibitem{belagiannis20153d}
Vasileios Belagiannis, Sikandar Amin, Mykhaylo Andriluka, Bernt Schiele, Nassir
  Navab, and Slobodan Ilic.
\newblock 3d pictorial structures revisited: Multiple human pose estimation.
\newblock {\em IEEE transactions on pattern analysis and machine intelligence},
  38(10):1929--1942, 2015.

\bibitem{benzine2020pandanet}
Abdallah Benzine, Florian Chabot, Bertrand Luvison, Quoc~Cuong Pham, and
  Catherine Achard.
\newblock Pandanet: Anchor-based single-shot multi-person 3d pose estimation.
\newblock In {\em Proceedings of the IEEE/CVF Conference on Computer Vision and
  Pattern Recognition}, pages 6856--6865, 2020.

\bibitem{bogo2016keep}
Federica Bogo, Angjoo Kanazawa, Christoph Lassner, Peter Gehler, Javier Romero,
  and Michael~J Black.
\newblock Keep it smpl: Automatic estimation of 3d human pose and shape from a
  single image.
\newblock In {\em European Conference on Computer Vision}, pages 561--578.
  Springer, 2016.

\bibitem{cao2018openpose}
Zhe Cao, Gines Hidalgo, Tomas Simon, Shih-En Wei, and Yaser Sheikh.
\newblock Openpose: realtime multi-person 2d pose estimation using part
  affinity fields.
\newblock {\em arXiv preprint arXiv:1812.08008}, 2018.

\bibitem{chen2020multi}
He Chen, Pengfei Guo, Pengfei Li, Gim~Hee Lee, and Gregory Chirikjian.
\newblock Multi-person 3d pose estimation in crowded scenes based on multi-view
  geometry.
\newblock {\em arXiv preprint arXiv:2007.10986}, 2020.

\bibitem{cheng2020higherhrnet}
Bowen Cheng, Bin Xiao, Jingdong Wang, Honghui Shi, Thomas~S Huang, and Lei
  Zhang.
\newblock Higherhrnet: Scale-aware representation learning for bottom-up human
  pose estimation.
\newblock In {\em Proceedings of the IEEE/CVF Conference on Computer Vision and
  Pattern Recognition}, pages 5386--5395, 2020.

\bibitem{corazza2010markerless}
Stefano Corazza, Lars M{\"u}ndermann, Emiliano Gambaretto, Giancarlo Ferrigno,
  and Thomas~P Andriacchi.
\newblock Markerless motion capture through visual hull, articulated icp and
  subject specific model generation.
\newblock {\em International journal of computer vision}, 87(1-2):156, 2010.

\bibitem{de2008performance}
Edilson De~Aguiar, Carsten Stoll, Christian Theobalt, Naveed Ahmed, Hans-Peter
  Seidel, and Sebastian Thrun.
\newblock Performance capture from sparse multi-view video.
\newblock In {\em ACM SIGGRAPH 2008 papers}, pages 1--10. 2008.

\bibitem{dong2019fast}
Junting Dong, Wen Jiang, Qixing Huang, Hujun Bao, and Xiaowei Zhou.
\newblock Fast and robust multi-person 3d pose estimation from multiple views.
\newblock In {\em Proceedings of the IEEE Conference on Computer Vision and
  Pattern Recognition}, pages 7792--7801, 2019.

\bibitem{elhayek2015efficient}
Ahmed Elhayek, Edilson de Aguiar, Arjun Jain, Jonathan Tompson, Leonid
  Pishchulin, Micha Andriluka, Chris Bregler, Bernt Schiele, and Christian
  Theobalt.
\newblock Efficient convnet-based marker-less motion capture in general scenes
  with a low number of cameras.
\newblock In {\em Proceedings of the IEEE Conference on Computer Vision and
  Pattern Recognition}, pages 3810--3818, 2015.

\bibitem{ershadi2018multiple}
Sara Ershadi-Nasab, Erfan Noury, Shohreh Kasaei, and Esmaeil Sanaei.
\newblock Multiple human 3d pose estimation from multiview images.
\newblock {\em Multimedia Tools and Applications}, 77(12):15573--15601, 2018.

\bibitem{fabbri2020compressed}
Matteo Fabbri, Fabio Lanzi, Simone Calderara, Stefano Alletto, and Rita
  Cucchiara.
\newblock Compressed volumetric heatmaps for multi-person 3d pose estimation.
\newblock In {\em Proceedings of the IEEE/CVF Conference on Computer Vision and
  Pattern Recognition}, pages 7204--7213, 2020.

\bibitem{fang2017rmpe}
Hao-Shu Fang, Shuqin Xie, Yu-Wing Tai, and Cewu Lu.
\newblock {RMPE}: Regional multi-person pose estimation.
\newblock In {\em ICCV}, 2017.

\bibitem{hartley2003multiple}
Richard Hartley and Andrew Zisserman.
\newblock {\em Multiple view geometry in computer vision}.
\newblock Cambridge university press, 2003.

\bibitem{he2017mask}
Kaiming He, Georgia Gkioxari, Piotr Doll{\'a}r, and Ross Girshick.
\newblock Mask r-cnn.
\newblock In {\em The IEEE International Conference on Computer Vision (ICCV)},
  2017.

\bibitem{he2020epipolar}
Yihui He, Rui Yan, Katerina Fragkiadaki, and Shoou-I Yu.
\newblock Epipolar transformer for multi-view human pose estimation.
\newblock In {\em Proceedings of the IEEE/CVF Conference on Computer Vision and
  Pattern Recognition Workshops}, pages 1036--1037, 2020.

\bibitem{huangend}
Congzhentao Huang, Shuai Jiang, Yang Li, Ziyue Zhang, Jason Traish, Chen Deng,
  Sam Ferguson, and Richard~Yi Da~Xu.
\newblock End-to-end dynamic matching network for multi-view multi-person 3d
  pose estimation.

\bibitem{huang2017towards}
Yinghao Huang, Federica Bogo, Christoph Lassner, Angjoo Kanazawa, Peter~V
  Gehler, Javier Romero, Ijaz Akhter, and Michael~J Black.
\newblock Towards accurate marker-less human shape and pose estimation over
  time.
\newblock In {\em 2017 international conference on 3D vision (3DV)}, pages
  421--430. IEEE, 2017.

\bibitem{iskakov2019learnable}
Karim Iskakov, Egor Burkov, Victor Lempitsky, and Yury Malkov.
\newblock Learnable triangulation of human pose.
\newblock In {\em Proceedings of the IEEE International Conference on Computer
  Vision}, pages 7718--7727, 2019.

\bibitem{jiang2020coherent}
Wen Jiang, Nikos Kolotouros, Georgios Pavlakos, Xiaowei Zhou, and Kostas
  Daniilidis.
\newblock Coherent reconstruction of multiple humans from a single image.
\newblock In {\em Proceedings of the IEEE/CVF Conference on Computer Vision and
  Pattern Recognition}, pages 5579--5588, 2020.

\bibitem{joo2015panoptic}
Hanbyul Joo, Hao Liu, Lei Tan, Lin Gui, Bart Nabbe, Iain Matthews, Takeo
  Kanade, Shohei Nobuhara, and Yaser Sheikh.
\newblock Panoptic studio: A massively multiview system for social motion
  capture.
\newblock In {\em Proceedings of the IEEE International Conference on Computer
  Vision}, pages 3334--3342, 2015.

\bibitem{joo2017panoptic}
Hanbyul Joo, Tomas Simon, Xulong Li, Hao Liu, Lei Tan, Lin Gui, Sean Banerjee,
  Timothy Godisart, Bart Nabbe, Iain Matthews, et~al.
\newblock Panoptic studio: A massively multiview system for social interaction
  capture.
\newblock {\em IEEE transactions on pattern analysis and machine intelligence},
  41(1):190--204, 2017.

\bibitem{joo2018total}
Hanbyul Joo, Tomas Simon, and Yaser Sheikh.
\newblock Total capture: A 3d deformation model for tracking faces, hands, and
  bodies.
\newblock In {\em Proceedings of the IEEE conference on computer vision and
  pattern recognition}, pages 8320--8329, 2018.

\bibitem{kanazawa2018end}
Angjoo Kanazawa, Michael~J Black, David~W Jacobs, and Jitendra Malik.
\newblock End-to-end recovery of human shape and pose.
\newblock In {\em Proceedings of the IEEE Conference on Computer Vision and
  Pattern Recognition}, pages 7122--7131, 2018.

\bibitem{kehl2006markerless}
Roland Kehl and Luc Van~Gool.
\newblock Markerless tracking of complex human motions from multiple views.
\newblock {\em Computer Vision and Image Understanding}, 104(2-3):190--209,
  2006.

\bibitem{li2019crowdpose}
Jiefeng Li, Can Wang, Hao Zhu, Yihuan Mao, Hao-Shu Fang, and Cewu Lu.
\newblock Crowdpose: Efficient crowded scenes pose estimation and a new
  benchmark.
\newblock In {\em Proceedings of the IEEE Conference on Computer Vision and
  Pattern Recognition}, pages 10863--10872, 2019.

\bibitem{liu2013markerless}
Yebin Liu, Juergen Gall, Carsten Stoll, Qionghai Dai, Hans-Peter Seidel, and
  Christian Theobalt.
\newblock Markerless motion capture of multiple characters using multiview
  image segmentation.
\newblock {\em IEEE transactions on pattern analysis and machine intelligence},
  35(11):2720--2735, 2013.

\bibitem{loper2015smpl}
Matthew Loper, Naureen Mahmood, Javier Romero, Gerard Pons-Moll, and Michael~J
  Black.
\newblock Smpl: A skinned multi-person linear model.
\newblock {\em ACM transactions on graphics (TOG)}, 34(6):1--16, 2015.

\bibitem{mahmood2019amass}
Naureen Mahmood, Nima Ghorbani, Nikolaus~F Troje, Gerard Pons-Moll, and
  Michael~J Black.
\newblock Amass: Archive of motion capture as surface shapes.
\newblock In {\em Proceedings of the IEEE International Conference on Computer
  Vision}, pages 5442--5451, 2019.

\bibitem{martinez2017simple}
Julieta Martinez, Rayat Hossain, Javier Romero, and James~J Little.
\newblock A simple yet effective baseline for 3d human pose estimation.
\newblock In {\em Proceedings of the IEEE International Conference on Computer
  Vision}, pages 2640--2649, 2017.

\bibitem{mehta2017vnect}
Dushyant Mehta, Srinath Sridhar, Oleksandr Sotnychenko, Helge Rhodin, Mohammad
  Shafiei, Hans-Peter Seidel, Weipeng Xu, Dan Casas, and Christian Theobalt.
\newblock Vnect: Real-time 3d human pose estimation with a single rgb camera.
\newblock {\em ACM Transactions on Graphics (TOG)}, 36(4):1--14, 2017.

\bibitem{newell2016associative}
Alejandro Newell, Zhiao Huang, and Jia Deng.
\newblock Associative embedding: End-to-end learning for joint detection and
  grouping.
\newblock {\em arXiv preprint arXiv:1611.05424}, 2016.

\bibitem{newell2016stacked}
Alejandro Newell, Kaiyu Yang, and Jia Deng.
\newblock Stacked hourglass networks for human pose estimation.
\newblock In {\em European conference on computer vision}, pages 483--499.
  Springer, 2016.

\bibitem{papandreou2018personlab}
George Papandreou, Tyler Zhu, Liang-Chieh Chen, Spyros Gidaris, Jonathan
  Tompson, and Kevin Murphy.
\newblock Personlab: Person pose estimation and instance segmentation with a
  bottom-up, part-based, geometric embedding model.
\newblock In {\em Proceedings of the European Conference on Computer Vision
  (ECCV)}, pages 269--286, 2018.

\bibitem{pavlakos2017harvesting}
Georgios Pavlakos, Xiaowei Zhou, Konstantinos~G Derpanis, and Kostas
  Daniilidis.
\newblock Harvesting multiple views for marker-less 3d human pose annotations.
\newblock In {\em Proceedings of the IEEE conference on computer vision and
  pattern recognition}, pages 6988--6997, 2017.

\bibitem{pishchulin2015deepcut}
Leonid Pishchulin, Eldar Insafutdinov, Siyu Tang, Bjoern Andres, Mykhaylo
  Andriluka, Peter Gehler, and Bernt Schiele.
\newblock Deepcut: Joint subset partition and labeling for multi person pose
  estimation.
\newblock {\em arXiv preprint arXiv:1511.06645}, 2015.

\bibitem{pishchulin2012articulated}
Leonid Pishchulin, Arjun Jain, Mykhaylo Andriluka, Thorsten Thorm{\"a}hlen, and
  Bernt Schiele.
\newblock Articulated people detection and pose estimation: Reshaping the
  future.
\newblock In {\em 2012 IEEE Conference on Computer Vision and Pattern
  Recognition}, pages 3178--3185. IEEE, 2012.

\bibitem{qiu2019cross}
Haibo Qiu, Chunyu Wang, Jingdong Wang, Naiyan Wang, and Wenjun Zeng.
\newblock Cross view fusion for 3d human pose estimation.
\newblock In {\em Proceedings of the IEEE International Conference on Computer
  Vision}, pages 4342--4351, 2019.

\bibitem{remelli2020lightweight}
Edoardo Remelli, Shangchen Han, Sina Honari, Pascal Fua, and Robert Wang.
\newblock Lightweight multi-view 3d pose estimation through camera-disentangled
  representation.
\newblock In {\em Proceedings of the IEEE/CVF Conference on Computer Vision and
  Pattern Recognition}, pages 6040--6049, 2020.

\bibitem{rogez2017lcr}
Gregory Rogez, Philippe Weinzaepfel, and Cordelia Schmid.
\newblock Lcr-net: Localization-classification-regression for human pose.
\newblock In {\em Proceedings of the IEEE Conference on Computer Vision and
  Pattern Recognition}, pages 3433--3441, 2017.

\bibitem{rogez2019lcr}
Gregory Rogez, Philippe Weinzaepfel, and Cordelia Schmid.
\newblock Lcr-net++: Multi-person 2d and 3d pose detection in natural images.
\newblock {\em IEEE transactions on pattern analysis and machine intelligence},
  42(5):1146--1161, 2019.

\bibitem{song2019end}
Jie Song, Bjoern Andres, Michael~J Black, Otmar Hilliges, and Siyu Tang.
\newblock End-to-end learning for graph decomposition.
\newblock In {\em Proceedings of the IEEE/CVF International Conference on
  Computer Vision}, pages 10093--10102, 2019.

\bibitem{song2020human}
Jie Song, Xu Chen, and Otmar Hilliges.
\newblock Human body model fitting by learned gradient descent.
\newblock In {\em Computer Vision--ECCV 2020: 16th European Conference,
  Glasgow, UK, August 23--28, 2020, Proceedings, Part XX 16}, pages 744--760.
  Springer, 2020.

\bibitem{song2017thin}
Jie Song, Limin Wang, Luc Van~Gool, and Otmar Hilliges.
\newblock Thin-slicing network: A deep structured model for pose estimation in
  videos.
\newblock In {\em Proceedings of the IEEE conference on computer vision and
  pattern recognition}, pages 4220--4229, 2017.

\bibitem{tu2020voxelpose}
Hanyue Tu, Chunyu Wang, and Wenjun Zeng.
\newblock Voxelpose: Towards multi-camera 3d human pose estimation in wild
  environment.
\newblock ECCV, 2020.

\bibitem{vlasic2008articulated}
Daniel Vlasic, Ilya Baran, Wojciech Matusik, and Jovan Popovi{\'c}.
\newblock Articulated mesh animation from multi-view silhouettes.
\newblock In {\em ACM SIGGRAPH 2008 papers}, pages 1--9. 2008.

\bibitem{wang2020hmor}
Can Wang, Jiefeng Li, Wentao Liu, Chen Qian, and Cewu Lu.
\newblock Hmor: Hierarchical multi-person ordinal relations for monocular
  multi-person 3d pose estimation.
\newblock In {\em European Conference on Computer Vision}, pages 242--259.
  Springer, 2020.

\bibitem{wang2020deep}
Jingdong Wang, Ke Sun, Tianheng Cheng, Borui Jiang, Chaorui Deng, Yang Zhao,
  Dong Liu, Yadong Mu, Mingkui Tan, Xinggang Wang, et~al.
\newblock Deep high-resolution representation learning for visual recognition.
\newblock {\em IEEE transactions on pattern analysis and machine intelligence},
  2020.

\bibitem{wei2016cpm}
Shih-En Wei, Varun Ramakrishna, Takeo Kanade, and Yaser Sheikh.
\newblock Convolutional pose machines.
\newblock In {\em CVPR}, pages 4724--4732, 2016.

\bibitem{wei2016convolutional}
Shih-En Wei, Varun Ramakrishna, Takeo Kanade, and Yaser Sheikh.
\newblock Convolutional pose machines.
\newblock In {\em Proceedings of the IEEE conference on Computer Vision and
  Pattern Recognition}, pages 4724--4732, 2016.

\bibitem{zanfir2018monocular}
Andrei Zanfir, Elisabeta Marinoiu, and Cristian Sminchisescu.
\newblock Monocular 3d pose and shape estimation of multiple people in natural
  scenes-the importance of multiple scene constraints.
\newblock In {\em Proceedings of the IEEE Conference on Computer Vision and
  Pattern Recognition}, pages 2148--2157, 2018.

\bibitem{zanfir2018deep}
Andrei Zanfir, Elisabeta Marinoiu, Mihai Zanfir, Alin-Ionut Popa, and Cristian
  Sminchisescu.
\newblock Deep network for the integrated 3d sensing of multiple people in
  natural images.
\newblock {\em Advances in Neural Information Processing Systems},
  31:8410--8419, 2018.

\bibitem{zhang20204d}
Yuxiang Zhang, Liang An, Tao Yu, Xiu Li, Kun Li, and Yebin Liu.
\newblock 4d association graph for realtime multi-person motion capture using
  multiple video cameras.
\newblock In {\em Proceedings of the IEEE/CVF Conference on Computer Vision and
  Pattern Recognition}, pages 1324--1333, 2020.

\bibitem{zhen2020smap}
Jianan Zhen, Qi Fang, Jiaming Sun, Wentao Liu, Wei Jiang, Hujun Bao, and
  Xiaowei Zhou.
\newblock Smap: Single-shot multi-person absolute 3d pose estimation.
\newblock In {\em European Conference on Computer Vision}, pages 550--566.
  Springer, 2020.

\end{thebibliography}
}

\end{document}